\documentclass{article}

\usepackage{microtype}
\usepackage{graphicx}
\usepackage{subfigure}
\usepackage{booktabs} % for professional tables
\usepackage{amsmath}
\usepackage[inline]{enumitem}
\usepackage{amsfonts}
\usepackage{verbatim}
\usepackage{booktabs}

\usepackage{hyperref}

\usepackage[accepted]{icml2021}

\icmltitlerunning{Curiosity-driven Exploration in Sparse-reward Multi-agent Reinforcement Learning}

\begin{document}
\newcommand{\gradient}{\nabla}
\twocolumn[
%\icmltitle{Improve Curiosity-driven with Go-Explore\\in Multi-agent Reinforcement Learning environment}
\icmltitle{Curiosity-driven Exploration in Sparse-reward Multi-agent Reinforcement Learning}


\icmlsetsymbol{equal}{*}

\begin{icmlauthorlist}
\icmlauthor{Jiong Li}{tue}
\icmlauthor{Pratik Gajane}{tue}
\end{icmlauthorlist}

\icmlaffiliation{tue}{Eindhoven University of Technology}
%\icmlaffiliation{goo}{Googol ShallowMind, New London, Michigan, USA}
%\icmlaffiliation{ed}{School of Computation, University of Edenborrow, Edenborrow, United Kingdom}

\icmlcorrespondingauthor{Jiong Li}{j.li11@student.tue.nl}
\icmlcorrespondingauthor{Pratik Gajane}{p.gajane@tue.nl}

\icmlkeywords{Machine Learning, ICML}
]



\printAffiliationsAndNotice{} % leave blank if no need to mention equal contribution

\begin{abstract}
%This document provides a basic paper template and submission guidelines.
%Abstracts must be a single paragraph, ideally between 4--6 sentences long.
%Gross violations will trigger corrections at the camera-ready phase.
Sparsity of rewards while applying a deep reinforcement learning method negatively affects %sustainability and 
its sample-efficiency. A viable solution to deal with the sparsity of rewards is to learn via intrinsic motivation which advocates for adding an intrinsic reward to the reward function to encourage the agent to explore the environment and expand the sample space. 
Though intrinsic motivation methods
%,such as intrinsic curiosity module and count-based method, 
are widely used to improve data-efficient learning in the reinforcement learning model, they also suffer from the so-called detachment problem.
%PG Explain detachment issues in a sentence (while trying to keep technical terms to a minimum)
%Meanwhile, another solution for the sparse reward issue has been proposed, so-called \textit{Go-Explore}, and this proposal shows its improvement in derailment and detachment from the curiosity-driven method. Though, Go-Explore suffers from the state simulation complexity issue. 
In this article, we discuss the limitations of intrinsic curiosity module in sparse-reward multi-agent reinforcement learning and propose a method called I-Go-Explore that combines the intrinsic curiosity module with the Go-Explore framework to alleviate the detachment problem.
\end{abstract}

\section{Introduction}
\label{sec:intro}
In a Reinforcement Learning (RL) task, an agent learns from the feedback given by the environment. In contrast to other paradigms of machine learning, an advantage of using reinforcement learning is the ability to solve complex learning tasks without expert domain knowledge. Particularly, in multi-agent reinforcement learning, the agent often operates in a complex environment. 
%such as a dynamic environment, a domain containing specialization knowledge, and so on. 
For instance, \citet{li2020cooperative} consider the problem of learning an optimal courier dispatching policy in a multi-agent dynamic environment.

A reinforcement learning problem is typically modeled using a Markov decision process (MDP) \citep{kaelbling1996reinforcement,sutton2018reinforcement}. An MDP is defined by a tuple that is formed by the state set, the action set, the transition function, and the reward function. The state set and the action set are typically assumed to be known to the agent while the transition function and the reward function are unknown to the agent. When an agent takes a particular action in a particular state, it receives a reward given by the reward function and the environment transitions to a state given by the transition function. 
%The state transition depends on the previous state and the chosen action. There is no fixed reward value as the goal in reinforcement learning, though the agent learns an optimal policy to maximize the reward signal. 
In a reinforcement learning problem, the learning goal is formalized as the outcome of
maximizing the obtained cumulative reward. In order to maximize the cumulative reward, a reinforcement learning agent faces a significant challenge known in the literature as the \textit{exploration-exploitation dilemma}. An agent may choose actions tried in the past and found to be rewarding (i.e. exploitation) or it may choose unexplored actions to see if they are more rewarding (i.e. exploration). Without sufficient exploration, an agent might not be able to find the optimal solution to the reinforcement learning problem. On the other hand, excessive exploration does not contribute to the goal of maximizing the cumulative reward and hence it represents sample-inefficiency. Thus, the ability to efficiently explore the environment remains key to sample-efficient reinforcement learning. 

Exploration in reinforcement learning is often driven via rewards and hence the density of rewards in the environment significantly influences the efficiency and thus sustainability of the model. 
The efficiency is presented by the training time of the model and the execution time to achieve certain goals. Sustainability measures the feasible range of the model.
%PG The connection to sustainability is not clear here and appears forced. Please clarify. 
%J I applied these keywords from the SNN topics to make a connection. This definition is my opinion.
Sparsity in the neural networks might improve the learning efficiency, however, a sparse-reward environment severely reduces the efficiency and sustainability of most reinforcement learning algorithms. Consequently, the sparsity of rewards diminishes the applicability of reinforcement learning to real-life problems as shown by \citet{wu2020joint}.%, they also mentioned that the traffic light controller also has a sparse reward issue.
%The sparse reward is one of the reinforcement learning challenges that has been studied for recent years. For instance, in \citep{wu2020joint}, they also mentioned that the traffic light controller also has a sparse reward issue.

%The algorithm is hard to converge to the optimization in a sparse reward environment since the algorithm optimizes the action policy to maximize the reward from the interaction with the environment. The sparsity of reward impedes the learning process of the algorithms and many solutions have come up to alleviate the side effect of the environment. 

A feasible solution to this problem is to improve the exploration efficiency to get as much reward as possible. For example, a solution known as \textit{reward shaping} uses some simulated positive rewards in the environment to encourage more exploration of the actual rewards from the interaction with the environment. However, reward shaping is sensitive to the reward density in a sparse-reward environment. 
Another possible solution is to apply intrinsic motivation techniques during the learning process.
Intrinsic motivation rewards the agent for exploring new states via an intrinsic reward.
As explained in \citet{andres2022evaluation}, there are four approaches to achieve intrinsic motivation -- \begin{enumerate*}[label=\arabic*)]
    \item \textit{count-based},
    \item \textit{prediction-error},
    \item \textit{random network distillation}, and 
    \item \textit{rewarding impact-driven exploration}.
\end{enumerate*} These methods are able to improve exploration in most cases, though they also have some drawbacks, such as \textit{derailment} and \textit{detachment}. 
Derailment describes a situation that the agent finds it hard to get back to the frontier exploration in the next episode since the intrinsic motivation rewards the seldom visited states.
When the intrinsic rewards run out during the exploration, the agent finishes the learning in current episode and goes back to the starting state for the next episode. However, the agent is expected to go further from the end position in the previous episode which is not attracted to the agent since some states are already visited in the last episode. This situation is known as detachment. 
%PG Here it feels like a sudden switch to ICM. Uptil now the discussion is about detachment and derailment and then a switch to ICM. Perhaps a connecting sentence would help$
The detachment and derailment issues are specially discussed in intrinsic motivation methods \citet{ecoffet2021first}.
Especially for the prediction-error approach, one of the popular implementations is via \textit{intrinsic curiosity module} (ICM). This is due to the ability of
ICM to deal with complex environments and domains \citep{yang2019flow, ladosz2022exploration, andres2022evaluation}. 
Accordingly, we use ICM as a baseline for our proposed solution.

This article focuses on improving the prediction-error intrinsic motivation method in the multi-agent reinforcement learning task to improve its sample-efficiency in sparse-reward environments.
In this work, we consider multi-agent game theoretic environment so as to simulate a complicated spares-reward environment which is close to reality. We focus on a mixed task which is not fully cooperative or not fully competitive. In the considered hybrid multi-agent environment, the proposed method can show the impact on exploration in cooperation as well as competition.
%within one environment. 
Each agent can only observe parts of the environment 
%known as partial observation Markov Game. 
and this partial observability increases the complexity of the problem. %Thus sample-efficient exploration for gathering environment information is crucial for effective learning.    
%The reinforcement learning algorithm is also crucial. 
In this article, we choose multi-agent deep deterministic policy gradient (MADDPG) as the learning algorithm with ICM. 
Since MADDPG allows each agent to apply different reward functions and supports the centralized training-decentralized execution framework, it is an efficient learning algorithm in this multi-agent environment.
We use ICM which provides the intrinsic reward function for each agent and the MADDPG algorithm allows the agent to get the external reward from the interaction with the environment. 
%In the multi-agent environment, the state transition is determined by the joint actions of all agents. Hence, the multi-agent reinforcement learning system might not be delighted with MDP. Therefore, the solutions for the single agent with a sparse reward environment have some improving space. 
%Improving the intrinsic motivation method shows a possibility of exploration effectively in reinforcement learning. T
%This article aims to show the exploration quality of the improvement of intrinsic motivation with a certain multi-agent reinforcement learning system and the exploration efficiency of the proposed method.  

% Contribution
%: Improve efficiency(avg reward, training time)
%: Discuss a possible improvement in multi-agent reinforcement learning environment.
%: combining exisiting method-> novel?
The main contributions of this work are as follows :
\begin{enumerate}
 \item We propose a novel solution to improve the training time and to find the optimal solution efficiently. 
 %PG : "to find the optimal solution efficiently". Is that fine? I do not want use the word policy because I do not think we have defined it yet.
    %and reward 
    %PG What do you mean by improving reward?
    %JL optimal the policy to gain the higher rewards
    Our solution also indicates a potential method to learn efficiently without using larger more complicated models.
    \item We analyze the drawbacks of some existing solutions in sparse-reward environment and compare the performance with the proposed method in a multi-agent environment.
    \item Based on the provided experiment results, we provide some future directions to further improve the proposed method paving the way for improving the efficiency of the reinforcement learning methods.
    \item Our work shows a way to deal with the sparsity and improve data-efficient learning in a sparse deep reinforcement learning environment by expanding the replay experience for an existing method.
\end{enumerate}

\subsection{Related Work}
\label{sec:related work}
The sparse-reward environment is commonly encountered in real-world applications, especially in multi-agent reinforcement learning problems. %The presence of sparse-rewards greatly limits the efficiency of reinforcement learning algorithms. 
Using intrinsic motivation shows a significant improvement in improving the sample efficiency in sparse-reward environments, for instance training a grandmaster in StarCraft \citep{vinyals2019grandmaster}, training the defensive escort team \citep{sheikh2020multi} etc. However, as \citet{delos2022curiosity} demonstrated, %the exploration result of the ICM method is not very ideal in some scenarios, 
exploration using ICM suffers from problems like detachment and inadequate exploration quality when used with limited observations. 

% ADD the same concern from other paper
\citet{burda2018exploration} also show that random network distillation (RND), another intrinsic motivated method, is sufficient to motivate the agent to explore efficiently in a short-term decision process, though it has a lower performance in a long-term goal. For instance, \citet{burda2018exploration} consider \textit{Montezuma’s Revenge}. Montezuma’s Revenge is an atari game in which the agent tries to collect useful items to escape with some actions, like jumping, running, sliding down poles, and climbing chains and ladders. However, the agent might get stuck if it focuses on short-term rewards (like collecting the items or fighting with enemies) but moves away from the exit door.
RND can help the agent to get the key, however, it does not realize that the key should be saved in a long-term strategy due to the limitation of the key. 
%There is the possibility to have sparse intrinsic rewards. 
Another solution is proposed in \citet{hester2018deep} which applied the pre-training with demonstration data to expand the replay buffer. The improved experiment results in the same environment, Montezuma’s Revenge, indicate that expanding the replay buffer might alleviate the detachment and derailment issue. However, collecting the domain knowledge and generating demonstration data requires extra knowledge of each experiment environment which might be unavailable or not conveniently attainable.
%PG Is the above statement correct?
%JL: I wanna explain that using domain knowledge might be inconvenient and hard to make genral utilization
%though this method motivated our work.

Go-Explore \citep{ecoffet2021first}
%PG for Go-Explore is there a published paper? The current reference is an ArXiV version. 
%J: I correct the cite with the nature version
is another method which aims to increase the efficiency of exploration. However, Go-Explore is more fragile in the multi-agent environment due to its complexity.
%as a solution for sparse-reward issue.
In this article, we propose to combine the features of Go-Explore and the ICM framework with the aim of solving detachment and derailment issues.

%%%%%%%%%%%%%%%%%%%%%%%%%%%%%%%%%%%%%%%%%%%%%%%%%%%%%%%%
\section{Problem Formulation}
%PG This section should be eiher be combined with the next section or renamed to something like "Research Questions" because it does not feature any problem formulation.
%PG : one solution for this issue is to move the description of multi-agent Markov game to this section as I have done now.

%In the face of the disadvantages mentioned in \ref{sec:related work}, the research questions we consider are as follows:
%\begin{enumerate}
    %\item In what cases that most likely to have sparse intrinsic rewards?
    %\item Which multi-agent environment most likely has detachment and derailment issue for the ICM method?
    %\item What is the performance(rewards and time cost) of the system only applied ICM method in this environment?
    %\item What is the  performance(average reward and time cost) of the agent combined with the ICM method and the Go-Explore method?
    %\item How far should the agents explore in the Go-Explore phases to access the best performance?
%\end{enumerate}

We design a decentralized partially observable environment involving multiple agents to simulate the real-world environment. The naive ICM method and the improved ICM method are implemented on a multi-agent deep deterministic policy gradient (MADDPG) algorithm.
%To simulate the general learning case, the Nash equilibrium will be implemented in the general sum game to optimize the agent's policy.
To quantify the improvement, we design a complicated sparse-reward environment in which a few specific states result in positive rewards. By comparing the total reward of our proposed method,  I-Go-Explore, with the total reward of the baseline naive ICM method, we showcase the performance improvement of the proposed method.

 We use the formulation of a partially observable Markov game to represent a multi-agent Markov game. A partially observable Markov game is described with the following quantities : 
\begin{itemize}
    \item $n$ agents.
    \item The public state space $\mathcal{S}$.
    \item The action space for each agent, denoted as $\mathcal{A}_1,...\mathcal{A}_n$.
    %PG Why is it {A}_n? The number of agents are N right? so it should be {A}_N or redefine  the number of agents as n.
    % J: since I used 'n' in the other subscription, I change "N agents" -> "n agents"
    \item The observation space for each agent, denoted as $\mathcal{O}_i,...,\mathcal{O}_n$.
    \item The stochastic policy for each agent as $\pi_i:\mathcal{O}_i\times\mathcal{A}_i\rightarrow[0,1]$. The next state is produced by the current state and joint action from each agent i.e., $\mathcal{T}:\mathcal{S}\times A_i \times...\times A_n\rightarrow\mathcal{S}$.
    \item The reward function for each agent is $r_i:\mathcal{S} \times \mathcal{A}_i \rightarrow \mathbb{R}$, then the agent $i$ arrives at a private observed state $o_i:\mathcal{S} \rightarrow \mathcal{O}_i$ and the private observation space is $\mathcal{O} = o_1,...,o_n$.
    %PG Something is undefined in the above item which causes the error of "Undefined control sequence." Please fix it
    \item The deterministic policies for each agent as $\mu_{\mathcal{O}_i}\rightarrow\mathcal{A}_i$ and the deterministic policy space is $\mu=\mu_1,...,\mu_n$.
    \item The policy parameters for each agent are $\theta_i$ and the policy parameters' space is $\Theta=\theta_1,...,\theta_n$. And the policy is represented by its parameters: $\pi_i\rightarrow\theta_i$.
    \item The action-state value function for each agent is calculated using the private observation space and the joint actions during the centralized training i.e., $Q_i^{\mu}:\mathcal{O}, a_1,...,a_n$ where $a_1\in \mathcal{A}_i,...a_n\in \mathcal{A}_n$. Each $Q_i^{\mu}$ is independent.     
\end{itemize}

%%%%%%%%%%%%%%%%%%%%%%%%%%%%%%%%%%%%%%%%%%%%%%%%%%%%%%%%

\section{Methodology}
In this section, we describe the methodology used in our article.
\subsection{Multi-Agent Deep Deterministic Policy Gradient (MADDPG)} 
\begin{figure}

\begin{center}
\includegraphics[width=\columnwidth]{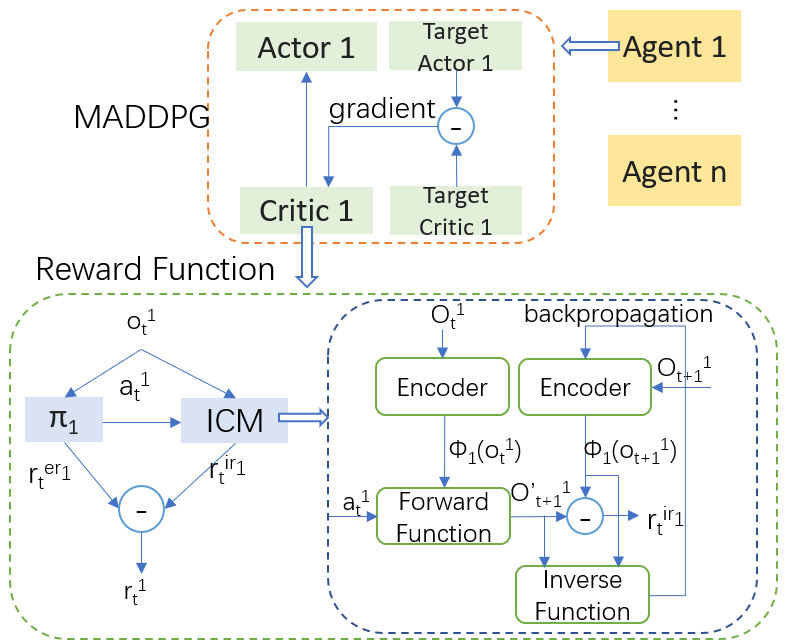}
\caption{MADDPG structure and intrinsic reward function for agent $1$. The critic network calculates the total reward signal for the chosen action and the total reward is the sum of the external reward and the intrinsic reward. %The structure of intrinsic curiosity module is shown below the figure to demonstrate the process of intrinsic reward.
}
\label{fig:maddpg}
\end{center}
\end{figure}

The MADDPG algorithm performs centralized training and decentralized execution to accelerate the training process using an actor-critic algorithm
\citep{NIPS1999_6449f44a}.
Since this algorithm is extended from Deep Deterministic Policy Gradient (DDPG), the parameters of the policy are optimized by the gradient descent method.
As for the actor-critic framework, the actor network achieves the decentralized execution which independently optimizes the policy only with self-observed information.
The actor network for each agent applies the policy gradient to maximize the expected reward signals and the optimization process of the policy is achieved by the gradient of the expected reward signals for each agent as:
\begin{align*}
    &\gradient_{\theta_i}\mathcal{J}(\theta_i) \nonumber \\
    &=\mathbb{E}_{\mathcal{O},a~D}[\gradient_{a_i}Q_i^{\mu}(\mathcal{O},a_1,...,a_n)\nabla_{\theta_i}\mu_i(o_i)|_{a_i=\mu_{\theta_i(o_i)}}] \label{eq:actor}
\end{align*}
%PG What is \nabla? Please define it using \newcommand
%PG Thanks for changing \nabla. I still get the error -- Undefined control sequence. Please look into it.
where $Q_i^{\mu}(\mathcal{O},a_1,...,a_n)$ is the decentralized action-value function for agent i, and $D$ is the experience replace buffer which contains the samples of trajectory at each episode.
As for the critic network for each agent, it achieves centralized training where it gathers all the information from each agent and evaluates the quality of the policy from the actor network. The loss function used in the critic network is:
\begin{align*}
    \mathcal{L}(\theta_i)&=\mathbb{E}_{\mathcal{O},a_1,...,a_n,r1,...,r_n,\mathcal{O}'}[Q_i^{\mu}(\mathcal{O},a_1,...,a_n)-y)^2] \nonumber \\
    y&=r_i+\gamma Q_i^{\mu'}(\mathcal{O}',a_1,...,a_n,r1,...,r_n)
\end{align*}
where y is the true reward signal from the next state and $\mu'$ is the target policy. However, in the general sum game, each agent will make a trade-off between the maximization of individual reward and the maximization of team reward.
%Furthermore, to make the policy optimization process more robust to the dynamic environment, \citep{lowe2017multi} suggested training a group of policies from K agent. At each episode, the collection randomly selected K sub-policy from K agent $\mu_{\theta_i}^{(k)}$. Instead of optimizing the individual reward signal, the actor network maximizes the reward signal of the group policies: 
% \begin{equation}\label{eq:maddpg}
%     \nabla_{\theta_i^{(K)}}\mathcal{J}(\theta_i)=\frac{1}{K}\mathbb{E}_{x,a~D}[\nabla_{a_i}Q_i^{\mu}(x,a_1,...,a_n)\nabla_{\theta_i^{(K)}}\mu_i^{(K)}(a_i|o_i)^{(K)}|_{a_i=\mu_i^{(K)}(o_i)}].
% \end{equation}
% where x is the current state space for the K agent. 
With this framework, centralized training and decentralized execution are implemented for the multi-agent Markov game. The reward function is designed based on the intrinsic motivation method to solve the sparse-reward issue. The structure of this framework is shown in Figure \ref{fig:maddpg} 
%PG the reference here says Figure 3.1 whereas the figure is labelled as Figure 1. Please look into it.
%PG this is fixed now. See the solution here - https://tex.stackexchange.com/questions/19651/reference-to-a-figure-uses-the-sections-number
% Thank you so muchXD
and further details are discussed below. The figure only illustrates the structure for a single agent and the other agents have the same structure.

\subsection{Intrinsic Curiosity Module}
%The ICM is one of the solutions for the sparse reward issue based on the intrinsic motivation concept\citep{pathak2017curiosity}. 
The ICM method \citep{pathak2017curiosity} motivates the agent to explore in multi-agent environment by giving an additional reward signal as an intrinsic reward. The intrinsic reward is the prediction error which is the difference between the predicted next state and the actual next state. The more disparate the prediction and the reality, the higher is the returned intrinsic reward.
For the decentralized execution, each agent has a private intrinsic reward module. ICM is composed of three parts: \textit{encoder}, \textit{forward model}, and \textit{inverse model}. 

Below we explain the intrinsic curiosity model for agent $i$.
%Taking the intrinsic curiosity model for agent i explains the detail of the components. 
The encoder abstracts the feature from the input state which is denoted as $\phi_i(o_i)$, where $o_i$ is the private observation for agent i. Let $f$ be a function  representing the forward model which generates a prediction of the next state with the current state feature and the action taken as input i.e.,
\begin{equation}\label{icm_nxt_o}
{o'}_{t+1}^i=f^i(o_t^i,a_t^i),
\end{equation}
where ${o'}_{t+1}^i$ is the prediction of the next observation of agent $i$ at time $t$.
The loss function used is the prediction error
\begin{equation}\label{icm_intrinsit_reward}
\mathcal{L}^i=\frac{1}{2}||{o'}_{t+1}^i-o_{t+1}^i||_2^2.
\end{equation}
The individual intrinsic reward is calculated using the following function $g$ which quantifies the novelty:
\begin{equation}\label{icm_nxt_o}
r_t^{ir_i}=g({o'}_{t+1}^i,o_{t+1}^i),
\end{equation}
where $r_t^{ir_i}$ is the intrinsic reward at time $t$ for agent $i$. The last part in the intrinsic curiosity module is the inverse model which uses the feature of current observation and the feature of next observation to predict a possible action at the current state. Then the inverse prediction error is used to train the inverse model and the encoder. The complete details of the ICM for agent $i$ are shown in Figure \ref{fig:maddpg}.

\paragraph{Reward Function in MADDPG}
The reward function is divided into two parts -- external reward signal and intrinsic reward signal. For agent $i$ at time $t$, the reward can be denoted as:
\begin{equation}\label{total_reward}
r_t^i=r_t^{ir_i}+r_t^{er_i},
\end{equation}
where $r_t^i$ is the total reward, $r_t^{ir_i}$ is the intrinsic reward and $r_t^{er_i}$ is the external reward. The intrinsic reward is generated using the ICM method. The external reward is decided by the environment. 
%From the motivation in the last section, the environment in this article simulates the general sum multi-agent system. 
Below we consider a function $h$ which quantifies the relationship between group $A$ and group $B$ where the agents in group $A$ are the opponents of the agents in group $B$. Similar to \citet{zhou2022multirobot}, the external reward signal is composed of the relative external reward and the base external reward. The relative external reward is calculated as:
\begin{equation}\label{external_reward}
r_t^{er_i}=\sum_{i=1}^{M}\sum_{j=1}^{N}h(A_i,B_j),
\end{equation}
where group $A$ contains $M$ agents and group $B$ contains $N$ agents. Furthermore, let $\delta$ be the base reward signal. Then, the sum of the two base reward signals of two groups is shown as:
%PG is the above statement correct?
%J: In our experiment case, sum(A,B) <= 0
\begin{align}
    r_t^{er_i}=\begin{cases}
    &r_t^{er_i}+\delta\quad \text{if agent belongs to group A}\\
    &r_t^{er_i}-\delta\quad \text{if agent belongs to group B}.
    \end{cases}
\end{align}\label{base_external_reward}
This reward function is used with the MADDPG method.

\subsection{I-Go-Explore}
%PG What does the title signify? Probably you want to say "Proposed solution : I-Go-explore"
Go-Explore \citep{ecoffet2021first}
starts the exploration from the achieved state which alleviates the problems of detachment and derailment for intrinsic motivation methods. The Go-Explore method has two phases: exploration and robustify. In the exploration phase, it first selects a state from the archive, simulates the chosen state, explores from the simulated state, and finally updates the exploration result to the archive. The archive is the visited state's space and  the selection is based on the probability distribution or a heuristic method.  As mentioned in \citet{ecoffet2021first}, detachment might happen when the intrinsic reward guides the agent to the frontier of the exploration and it loses interest in going back to the exploration state. Therefore, we propose a solution to add an exploration phase at the end of each episode to give additional motivation so that the experience in the replay buffer is expanded. This exploration is decentralized and each agent explores with fixed steps. As for the simulation step, the trajectory, the action sequence, and the reward signal also need to be replayed
%PG replay -> replayed?
from the storage. The newly visited states will be updated to the achievement of agent i. The existing state will be updated with fewer trajectory steps in the achievement. In our experiments, we show that the performance and training efficiency of reinforcement learning can be improved  by our proposed solution.

\section{Experimental Results}

This article is extended based on the success of the ICM method in multi-agent reinforcement learning sparse reward environment and the success of Go-Explore in a single-agent sparse reward environment. The initial multi-agent reinforcement learning (MARL) environment implements \textit{Multi-agent Particle Environment} (MPE) which is proposed in \citet{lowe2017multi}. This environment was created for MADDPG method and it contains several scenarios to test an algorithm's performance in a cooperative task \footnote{The code used for the experiments will be made available via a Github repository on acceptance of this submission.}.
% PG What is the reason for adding the above paragraph? 
% J: it's a background of the experiment environment and motivates the framework's parameter setting
%In our work, we take \textit{Predator-prey} scenario as the experiment environment which contains 1 prey, 3 predators, and 2 landmarks as the obstacles. An  illustration of this environment is shown in Figure \ref{fig:prey}.
%
\begin{figure}
\centering
\includegraphics[height=6.5cm]{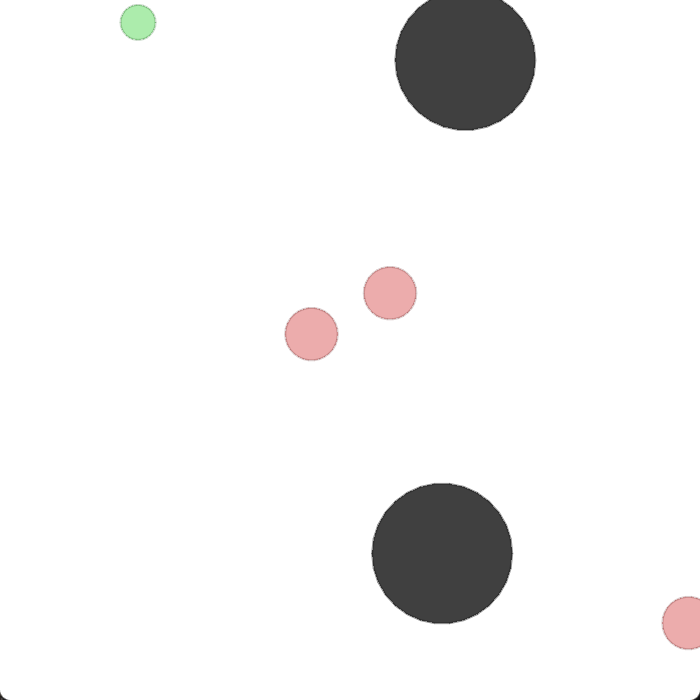}
\caption{The large black circles represent the landmark, the red circles are the predators and the green circle is the prey.}
\label{fig:prey}
\end{figure}

In the given experiments, the base model applies MADDPG as the learning algorithm so that each agent has an individual reward function.
Based on MADDPG framework, we test the performance of our proposed solution  I-Go-Explore against the baseline of ICM.

In the MADDPG framework, each agent has its own actor-critic network and target actor-critic network which contains two-layer ReLU MLPs with 64 cells in each layer. The actor-critic network updates the parameters in each episode and the target actor-critic network applies the soft update function to optimize the parameters in a short time period. Usually, the learning rate of the critic network is slightly higher than the actor network's since the aim of the critic network is to evaluate the actor network. In the ICM method, the encoder has two ReLU MLPs with 64 cells in each layer, and the forward function and inverse both have one ReLU MLP with 64 cells in each layer. In the experiment, we applied the same environment as \citet{lowe2017multi}. Therefore, the basic parameters setup are following \citet{lowe2017multi}, except updating frequency for target network since the aim of this paper focuses on the performances of intrinsic motivation instead of optimizing MADDPG framework. The target network is updated every 10 steps during the training to alleviate the side effects of the partial observation environment.

We conduct three sets of experiments to show that our proposed solution provides a promising improvement in the multi-agent environment. There are three predators, one prey and two landmarks (obstacles) in this chasing environment. The prey moves faster than the predator. The prey is punished for being caught or moving out of the vision border which is -1.0 to +1.0. The prey is easy to be caught by moving out of the border.
%The predators are rewarded by catching the prey and are not punished by moving out of the vision border.
%PG What about the predators? How are they rewarded?
%I see that this is getting repeated later, so I am just moving that part here. 
 The prey gets a punishment of -10 for each occurrence of the above two situations. The predators get rewards by catching the prey with +10 for each catch and they are not punished for moving out of the vision border.
% PG could you also mention the value of the rewards? Like is it +1 for a specific action?
An illustration of this environment is shown in Figure \ref{fig:prey}.

 Each agent has an individual policy and a reward system to maximize the individual reward. A more optimal solution is for the predators to learn to cooperate to chase the prey which indicates the maximization of the total reward. The prey-predator is a two-dimensional environment and the state is presented by the positions of agents and landmarks, and the action is formed by five discrete moving directions. However, each agent has an observation limitation that only allows the agent to observe the partial environment within a circular area of radius 0.5.
 
The aim of the first experiment is to observe the performances of both methods in terms of the reward of prey and predators and also the time-consumption.
For the second experiment, we aim to compare the long-term training result with the short-term training result to analyze if I-Go-Explore 
%PG I find the term "I-Go-Explore ICM" slighly strange. Could we use something else? 
% J: Please help me name this method XD...
%PG A catchy acronym (for example, I had proposed an algorithm with the name MNM which sounds like M&M chocolates) would be the best. Otherwise just go for Go-explore+ICM
% J: lol... GEIC? Or let's take Go-explore+ICM
indicates a good effect on the long-term task. The third experiment focuses on the I-Go-Explore method which tests the performances with different exploration lengths to observe the impact of exploration length. 
The evaluation setting during the training for the three experiments is the same. We evaluate the rewards of each agent for every 5 episodes since the environment is stochastic and it would be more reasonable to evaluate average rewards for a batch of training episodes. 
There are training phases and test phases for the experiments. We trained the model according to the experiment settings and applied the trained model to the corresponding test environment.
%to observe.
%In the experiment, the aim is to show that the ICM method might have a detachmenissue in the sparse reward environment.
%In this section, we focus on showing the experiments' results and analyzing them. In the end, we summarize the experiments and list the future work based on the experiment results.

\subsection{Experiment 1}
\begin{figure}
\begin{center}
\includegraphics[width=\columnwidth]{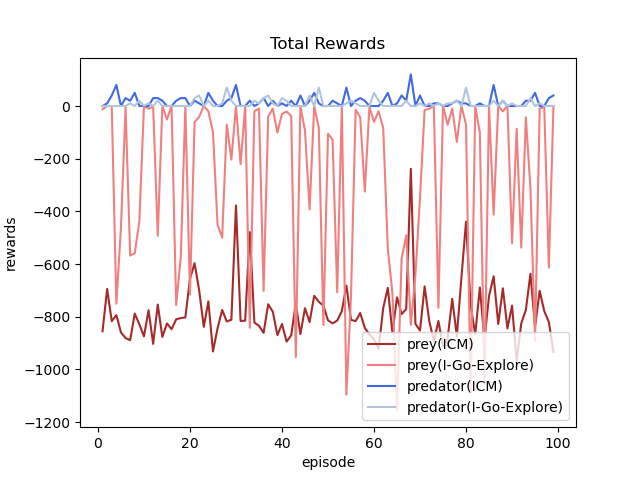}
\caption{The figure compares the reward of prey and predators for 500 episodes with 100 steps in ICM method and I-Go-Explore method.}
\label{fig:exp1}
\end{center}
\end{figure}

The aim of the first experiment is to observe the performances of both methods in terms of the reward of prey and predators and also the time-consumption. We trained the model for ICM and I-Go-Explore separately for 500 episodes with 100 steps in each episode. Then, we test the rewards of ICM model and the I-Go-Explore model in the prey-predator environment (different setting from the training) in 100 episodes with 100 steps in each episode. With the same parameter setups for the training process in ICM and I-Go-Explore, I-Go-Explore has higher average rewards during the test as shown in Figure \ref{fig:exp1}. We took 50 steps (half the length of the episode) for the exploration phase in I-Go-Explore method since the exploration information might help the sampling during the target network training and we do not want the impact of the exploration to be stronger than the actual training experience.
%PG Do not use placement options like [ht] with figures or other floating environments. Let LaTeX handle the placement. 
%J: I will fix the rest

%PG I think there needs to be some restructuring of  this section. Currently it is description of experiment 1, experiment 2 and experiment 3 together. And then results of experiment 1, experiment 2 and experiment 3 together. I think it will be easier to read if it is description of experiment 1 and then its result, then description experiment 2 and and its result followed by description experiment 3 and then its result. Could you please restructure it this way?
Nevertheless, from the results shown in Figure \ref{fig:exp1}, some outliers catch our attention, like the episode around 55 and the episode around 65. Recall that the prey is punished for getting caught while, predators are rewarded for catching the prey. These outliers indicate that the prey is easily caught by the predators in those episodes and policies for the predators are better. To confirm our conjecture, we checked the images
%PG gif images -> images?
in this evaluation period and each image records the moves of each agent during the episode. Observing the image with the large negative prey reward shows that the prey always runs out of the borders to escape the predators and predators successfully move out of the borders to catch the prey. In the I-Go-Explore method, the prey and predators got 0 rewards. One possible explanation is that the prey has a larger velocity and it learns to escape the predators and keeps itself within the border. On the other hand, the predators are too slow to catch the prey even though they know how to move toward the prey.

The average rewards for prey and predators under each method are shown in Table \ref{tab:exp1}. From the average rewards, 
%among 500 episodes result, 
we observe that prey can avoid being caught while keeping itself within the borders in I-Go-Explore method. We conjecture that it benefited from the exploration so that the MADDPG model gave better guidance during the test.
%PG Could you rephrase this too? it is slightly unclear as to what the extra environment information is

\begin{table}
\centering
\begin{tabular}{ c c c}
 \toprule
agent    & ICM   & I-Go-Explore \\ 
\midrule
prey     & -786.62 & -268.83         \\ 
predator (each) & 16.76     & 8.88               \\ 
 \bottomrule
\end{tabular}
\caption{The average rewards for 500 episodes in two methods are shown in here.}
\label{tab:exp1}
\end{table}
We trained the model on Intel Core i5-8265 cpu with 8GB RAM and the go-explore ICM method took less training time (15805 seconds) than the ICM method (18005 seconds).
%PG If you are mentioning the time taken, you should mention the details about the computer on which you executed the code like CPU capabilities and memory.
%PG Please do not manipulate space by using vspace or vskip anywhere. That is considered a violation of the submission requirements. 

\subsection{Experiment 2}
\begin{figure}
\begin{center}
\includegraphics[width=\columnwidth]{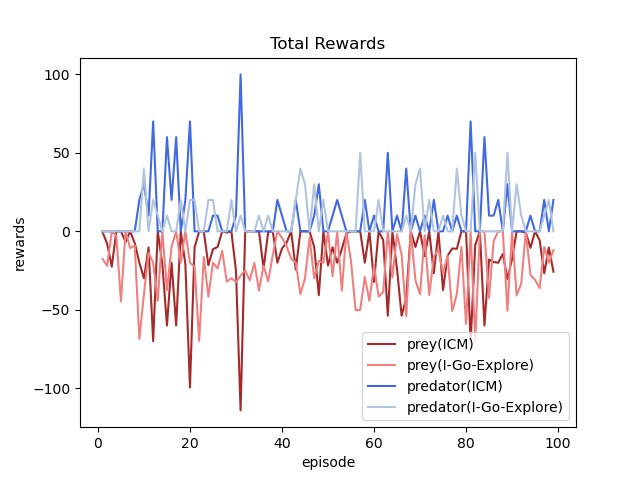}
\caption{The figure is the intrinsic reward for each agent. Agent 4 is the prey and the rest are the predators.}
\label{fig:exp2}
\end{center}
\end{figure}

To compare the impact of I-Go-Explore on long-term tasks from experiment 1, the training step in experiment 2 is 20 steps for each episode. Similar to the  exploration performed in experiment 1, the exploration phase takes half the episode length which is 10 steps for go-explore ICM method. The total reward in two methods for 100 episodes with 20 steps is plotted in Figure \ref{fig:exp2}. From the results, there is little difference between the two models in the test environment. Indeed, in the shorter steps of training, ICM method showed better performance in prey agent which is the opposite of longer steps of training. The average rewards of the short-term training and the long-term training for two methods are shown in Table  \ref{tab:exp2}. Also, there is little difference between the computation time of the two methods.

\begin{table}[]
\centering
\begin{tabular}{c c c c c }
\toprule
agent & \begin{tabular}[c]{c@{}} ICM\\(long) \end{tabular}                &\begin{tabular}[c]{@{}c@{}} I-Go\\-Explore\\(long) 
   \end{tabular}
      &\begin{tabular}[c]{c@{}} ICM\\(short) \end{tabular}
      &\begin{tabular}[c]{@{}c@{}} I-Go\\-Explore\\(short) \end{tabular}\\ 
      \midrule
\multicolumn{1}{ c }{prey} & \multicolumn{1}{c }{-786.62} & \multicolumn{1}{c}{-268.83} & \multicolumn{1}{c }{-15.66} & \multicolumn{1}{c }{-22.96} \\ 
predator                   & 16.76                         & 8.88                         & 10.70                       & 7.77                        \\ 
\bottomrule
\end{tabular}
\caption{This table compares the difference of the rewards between two methods in long-term training and short-term training.}
\label{tab:exp2}
\end{table}

%PG Table {tab:exp2} is going outside the column. Please resolve this problem.

\subsection{Experiment 3}
\begin{figure}
\begin{center}
\includegraphics[width=\columnwidth]{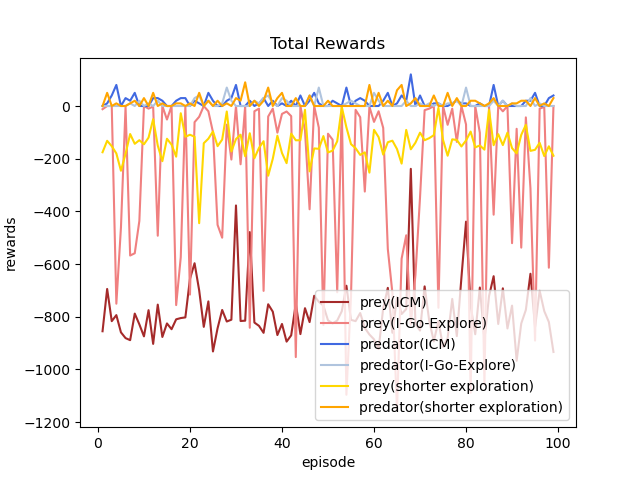}
\caption{The red line represents the total reward value for predators and the green represents the total reward value for prey}
\label{fig:exp3}
\end{center}
\end{figure}

In the last experiment, we compared the I-Go-Explore method with different exploration settings in 100 episodes of 100 steps each. The short-exploration model is trained with 10 steps in the exploration phase and the long-exploration model is trained with 50 steps in the exploration phase. The framework parameter settings are the same for both the models except for the exploration length. In Figure  \ref{fig:exp3}, we compared ICM, I-Go-Explore with 50 steps exploration, and I-Go-Explore with 10 steps exploration. From the experimental results, for prey agent, go-explore ICM model with a shorter exploration phase has a better test performance as shown in Table \ref{tab:exp3}. One possible explanation is that too much arbitrary exploration might have a side effect on the policy in this environment.

\begin{table}[]
\centering
\begin{tabular}{ c c c c }
\toprule
agent    & ICM     &\begin{tabular}[c]{c@{}} I-Go-Explore\\ ICM(long) 
   \end{tabular}&\begin{tabular}[c]{c@{}} I-Go-Explore\\ ICM(short) \end{tabular}\\ 
   \midrule
prey     & -786.62 & -268.83               & -143.10                \\ 
predator & 16.76    & 8.88                  & 14.14                  \\ 
\bottomrule
\end{tabular}
\caption{This table compares the I-Go-Explore method trained with different exploration lengths and the shorter exploration length gets a better result in the experiment.}
\label{tab:exp3}
\end{table}

\section{Concluding Remarks and Future Work}

In this article, we showed that the proposed method I-Go-Explore shows promising results for the task of improving the performance of the existing reinforcement learning methods in a data-sparse environment. We also showed how the I-Go-Explore method uses sparsity to improve the model's efficiency. This article indicates a promising research direction to improve the model efficiency with sparsity.

Our experimental results showcase the better performance of I-Go-Explore compared to the baseline of the ICM method. 
%Nevertheless, these experiments are insufficient to prove the success of the I-Go-Explore method. 
%There are some future works from several aspects as follows:
Interesting directions for future work include the following. 
\begin{itemize}
    \item For learning algorithm : ICM method has been implemented with some other algorithms, like COMA \citep{delos2022curiosity,yang2021ciexplore}, PPO \citep{zhang2022proximal}. 
    %To increase the credibility of the proposed method, 
    We could implement I-Go-Explore with such different learning algorithms.
    \item %For the experiment environment, reinforcement-learning results are usually affected by the environment. 
    To show the robustness of the I-Go-Explore method, we could test more partially observable multi-agent environments.
    %\item For the experiment setup, due to the time limitation, the test is quite simple. And the experiment results might be slightly influenced by the random initial environment setup. In the future, we should strictly control the factors in the environment and compare the performance in the test environment after fully the complete training and testing the models in a more complicated environment.
\end{itemize}

%From the experiment plots, it's able observed that the intrinsic reward has a declining trend during the training. However, the exploration might insufficient since the total reward of the predators still gets a negative result. Obviously, the policy of each predator is not very optimal which is influenced by the exploration experience. Different environment setting also influences the quality of the ICM method and the same effect for the different parameter settings of MADDPG algorithm \citep{andres2022evaluation}. The experiment in is only to get an insight into the proposed problem and show a possible improvement of existing method. In the later research phases, I will construct more environments with the proposed method.

% \begin{figure}[ht]
% \vskip 0.2in
% \begin{center}
% \centerline{\includegraphics[width=\columnwidth]{icml_numpapers}}
% \caption{Historical locations and number of accepted papers for International
% Machine Learning Conferences (ICML 1993 -- ICML 2008) and International
% Workshops on Machine Learning (ML 1988 -- ML 1992). At the time this figure was
% produced, the number of accepted papers for ICML 2008 was unknown and instead
% estimated.}
% \label{icml-historical}
% \end{center}
% \vskip -0.2in
% \end{figure}

\bibliography{main}
\bibliographystyle{icml2021}

\end{document}